\documentclass{article}
\usepackage{ijcai22}

\usepackage{times}

\usepackage{soul}
\usepackage{url}
\usepackage[hidelinks]{hyperref}
\usepackage[utf8]{inputenc}
\usepackage[small]{caption}
\usepackage{graphicx}
\usepackage{amsmath}
\usepackage{booktabs}
\newtheorem{theorem}{Theorem}
\usepackage{subcaption}  % subfigure

\usepackage{amssymb}
\usepackage{pifont}
\usepackage{xcolor}
\newcommand{\xmark}{\ding{55}}%

\title{Improved Robustness of Vision Transformer \\ via PreLayerNorm in Patch Embedding}

\author{
Bum Jun Kim$^1$\and
Hyeyeon Choi$^1$\and
Hyeonah Jang$^1$\and
Dong Gu Lee$^1$\and\\
Wonseok Jeong$^2$\and
Sang Woo Kim$^{1,2}$\footnote{Contact Author}\\
\affiliations
$^1$Department of Electrical Engineering, Pohang University of Science and Technology\\
$^2$Graduate School of Artificial Intelligence, Pohang University of Science and Technology\\
\emails
\{kmbmjn, hyeyeon, hajang, dgleee, wonseok.jeong, swkim\}@postech.edu
}

\begin{document}

\maketitle

\begin{abstract}
Vision transformers (ViTs) have recently demonstrated state-of-the-art performance in a variety of vision tasks, replacing convolutional neural networks (CNNs). Meanwhile, since ViT has a different architecture than CNN, it may behave differently. To investigate the reliability of ViT, this paper studies the behavior and robustness of ViT. We compared the robustness of CNN and ViT by assuming various image corruptions that may appear in practical vision tasks. We confirmed that for most image transformations, ViT showed robustness comparable to CNN or more improved. However, for contrast enhancement, severe performance degradations were consistently observed in ViT. From a detailed analysis, we identified a potential problem: positional embedding in ViT's patch embedding could work improperly when the color scale changes. Here we claim the use of PreLayerNorm, a modified patch embedding structure to ensure scale-invariant behavior of ViT. ViT with PreLayerNorm showed improved robustness in various corruptions including contrast-varying environments.
\end{abstract}

\section{Introduction}
\label{sec:introduction}

Recently, vision transformers (ViTs) have shown state-of-the-art performance in various vision tasks, replacing the domain of convolutional neural networks (CNNs) \cite{dosovitskiy2020image}. Here, CNN uses operations such as convolution, pooling, and non-linearity to combine and compress image features. In contrast, ViT divides an image into patches and models the sequence of patches with a transformer composed of self-attention and multi-layer perceptron (MLP).

Because of this distinct architecture, ViT can exhibit different behavior from CNN. Since the inner behavior of ViT is still not much known, more research is needed. For example, the original paper on ViT conjectured that ViT has weaker translation invariance than CNN. If a 1-pixel shifted image leads to a completely different result in a practical vision system, the vision system will be considered unreliable. Also, suppose that an autonomous vehicle is equipped with a vision system that is not robust to brightness changes. Then, a brightness change may cause an unexpected driving decision, resulting in a fatal accident. As such, in the practical vision task, it is important to understand and improve the robustness of the model for various changes in the image \cite{ford2019adversarial,yin2019fourier,hendrycks2019augmix}.

This paper investigates the behavior and robustness of ViT and CNN. First, the comprehensive verification results are presented to see a difference in robustness between ViT and CNN due to the architectural difference (Section \ref{sec:robustnesstest}). The performance of ViT and CNN are quantified in various image corruptions including translation, rotation, brightness, and contrast. According to our robustness test benchmark, for most corruptions, ViT showed robustness comparable to or stronger than CNN. This observation is consistent with the findings of other studies evaluating robustness \cite{dosovitskiy2020image,naseer2021intriguing}. However, we discovered that ViT is particularly vulnerable to contrast-varying environments. The weakness of ViT for contrast-enhanced images was consistently observed for various datasets and variants of ViT. We emphasize that the difference in robustness came from the difference in architectures since the same training environment and data augmentation rule are used.

Therefore, ViT's intrinsic architecture is expected to have certain vulnerabilities different from CNN. The cause we argue lies in the patch embedding in ViT. We will prove that because positional embedding in patch embedding serves as a fixed bias, ViT does not respond flexibly to a change in a color scale. We will show that PreLayerNorm, which has been optionally used in some other variants of ViT, plays an important role in ensuring the consistent behavior of positional embedding from the changes in color scale. We confirmed that ViT with PreLayerNorm showed improved robustness in the same contrast test (Section \ref{sec:detailedanalysisonpatchembedding}).

\section{Robustness Test}
\label{sec:robustnesstest}
In this section, we compare the robustness of ViT and CNN for various image corruptions.

We target four popular CNNs and two ViTs: ResNet-50 \cite{he2016deep}, Wide-ResNet-50-2 \cite{zagoruyko2016wide}, ResNeXt-50-32x4d \cite{xie2017aggregated}, EffNetV2-L \cite{tan2021efficientnetv2}, ViT-L \cite{dosovitskiy2020image}, and Swin-L \cite{liu2021swin}. Using \texttt{PyTorch Image Models} \cite{rw2019timm,paszke2019pytorch}, ImageNet pre-trained weights were obtained to perform fine-tuning. Fine-tuning was performed on two datasets, Stanford Cars \cite{krause20133d} and Stanford Dogs \cite{khosla2011novel}. All experiments are conducted for $224\times224$ resolution using the standard data augmentation such as random resized crop and color jitter (See Appendix). For training, stochastic gradient descent with momentum 0.9 \cite{sutskever2013importance}, learning rate \{0.01, 0.001\}, cosine annealing schedule with 200 iterations \cite{loshchilov2016sgdr}, weight decay 0.0005, batch size 32 were used. The dataset was split into train, val, and test sets at a ratio of 70:15:15. A model with the best validation accuracy was obtained in 200 epochs training. Note that since our goal is to observe differences in robustness due to differences in architecture, we set the same training environment for both CNN and ViT. However, for ViTs, a different training environment (e.g., higher weight decay) may help improve performance \cite{steiner2021train}. Therefore, in the results below, there is room where the performance of ViTs can be further improved. However, rather than improving the performance, we focus on observing the difference in robustness owing to the different architecture.

For each obtained model, we applied image transformations such as translation, rotation, brightness, and contrast to the test dataset and measured the test accuracy for each corruption and severity. In summary, several CNNs and ViTs are trained in the same environment and applied to the same corruption. Through the difference in test accuracy resulting from the difference in intrinsic architecture, the robustness of each architecture is quantitatively evaluated.

\subsection{Translation Test}
\label{sec:translationtest}
First, we assumed a situation in which translation is applied to an image. In particular, the original paper on ViT conjectured that ViT has weaker translation invariance than CNN, which needs to be verified experimentally.

Previous studies on translation invariance have been theoretical because they shifted the zero-padded window or measured top probability rather than accuracy \cite{zhang2019making,azulay2018deep}. We consider a more practical scenario. In standard data augmentation, the test image is resized so that the short side length becomes 256 pixels to obtain a template, and then the centered window $224 \times 224$ is cropped. Assuming a camera shift scenario, when setting the window in the template, we shifted the window center coordinates by $+s$ pixels horizontally and vertically and then cropped it. This protocol is a more realistic scenario in that it allows the natural background without zero-padding the edge area.

\begin{figure*}
  \centering
  \begin{subfigure}{0.35\linewidth}
    \includegraphics[width=0.99\linewidth]{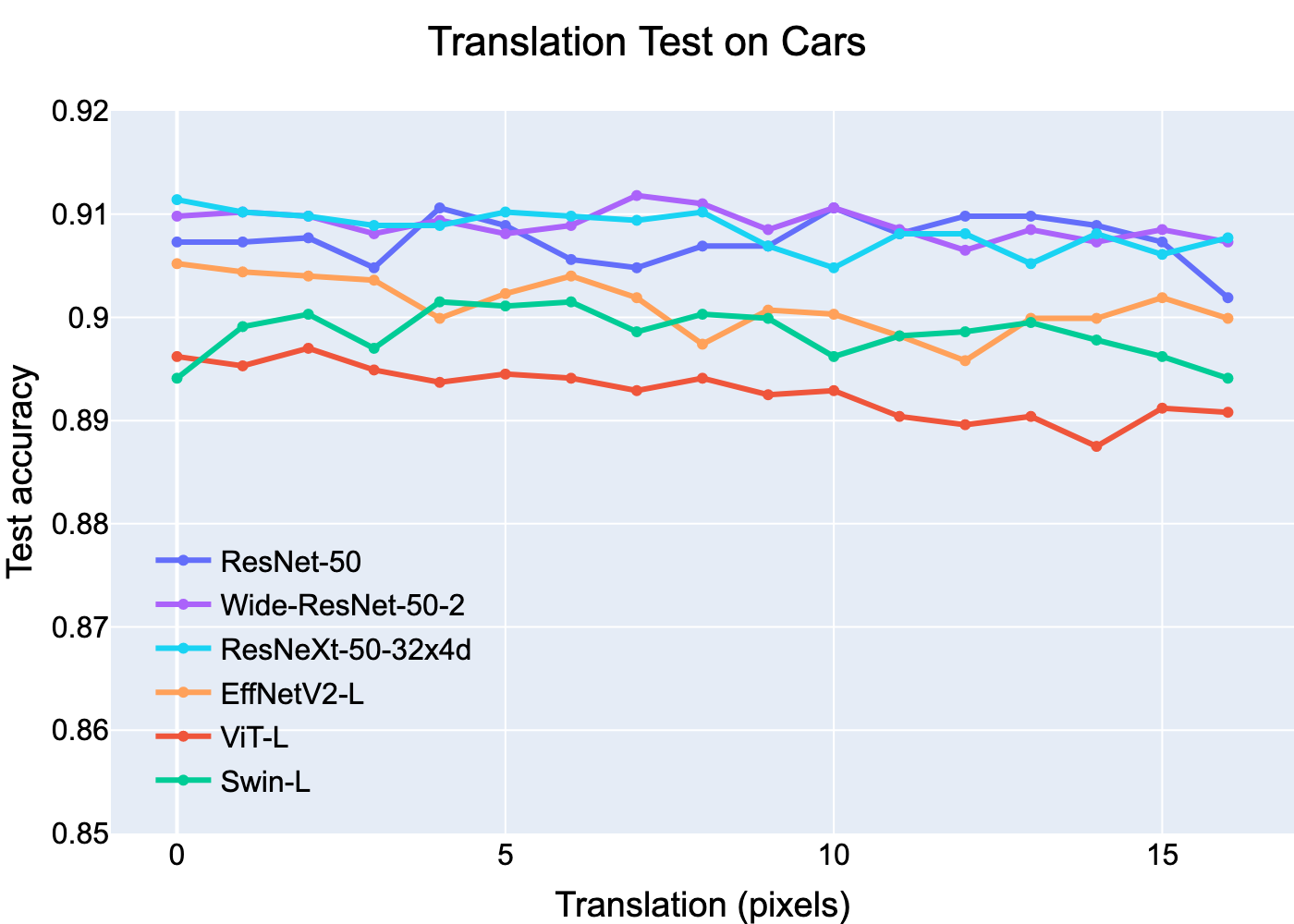}
  \end{subfigure}
  \hfill
  \begin{subfigure}{0.35\linewidth}
    \includegraphics[width=0.99\linewidth]{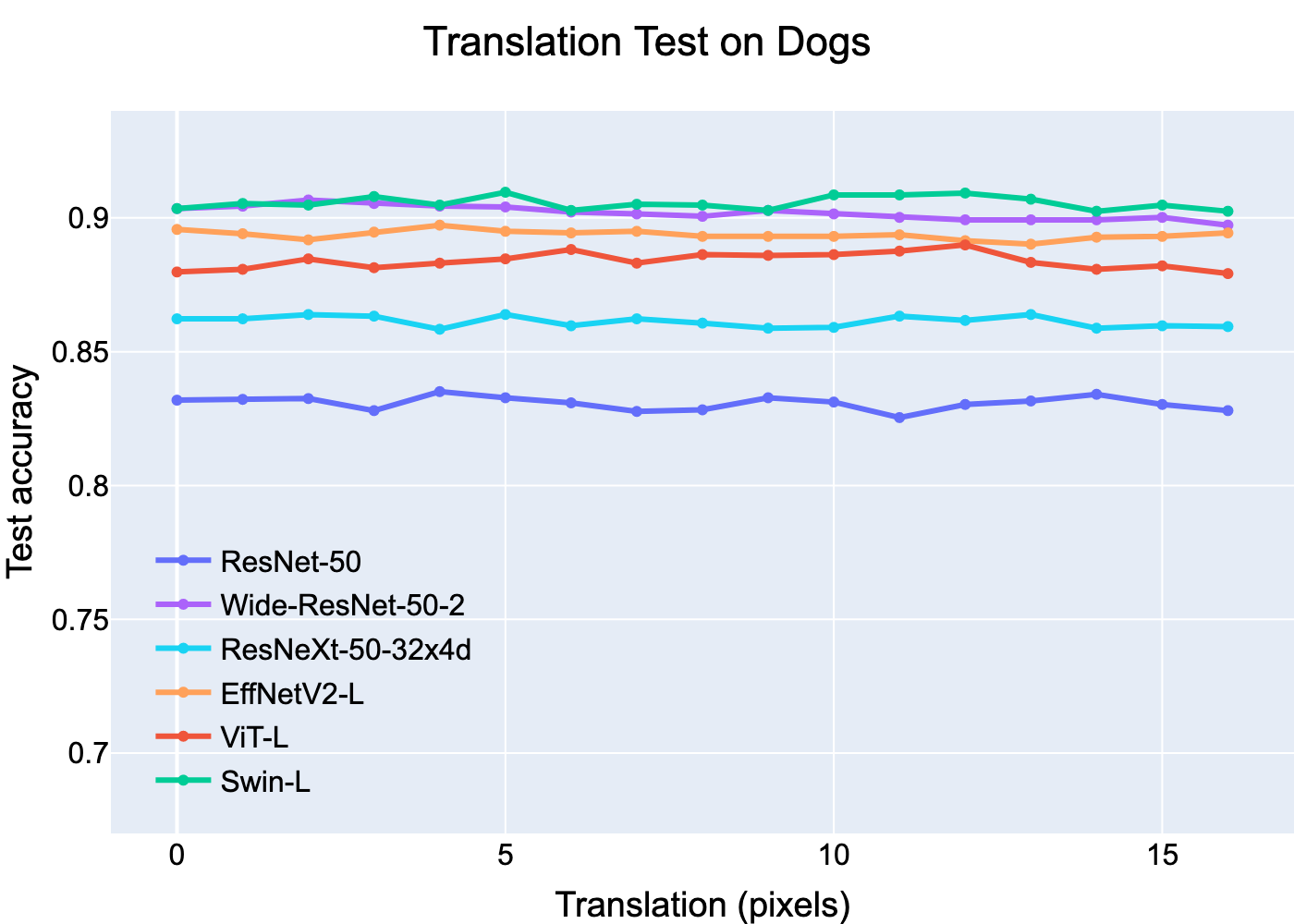}
  \end{subfigure}
  \hfill
  \begin{subfigure}{0.29\linewidth}
    \includegraphics[width=0.99\linewidth]{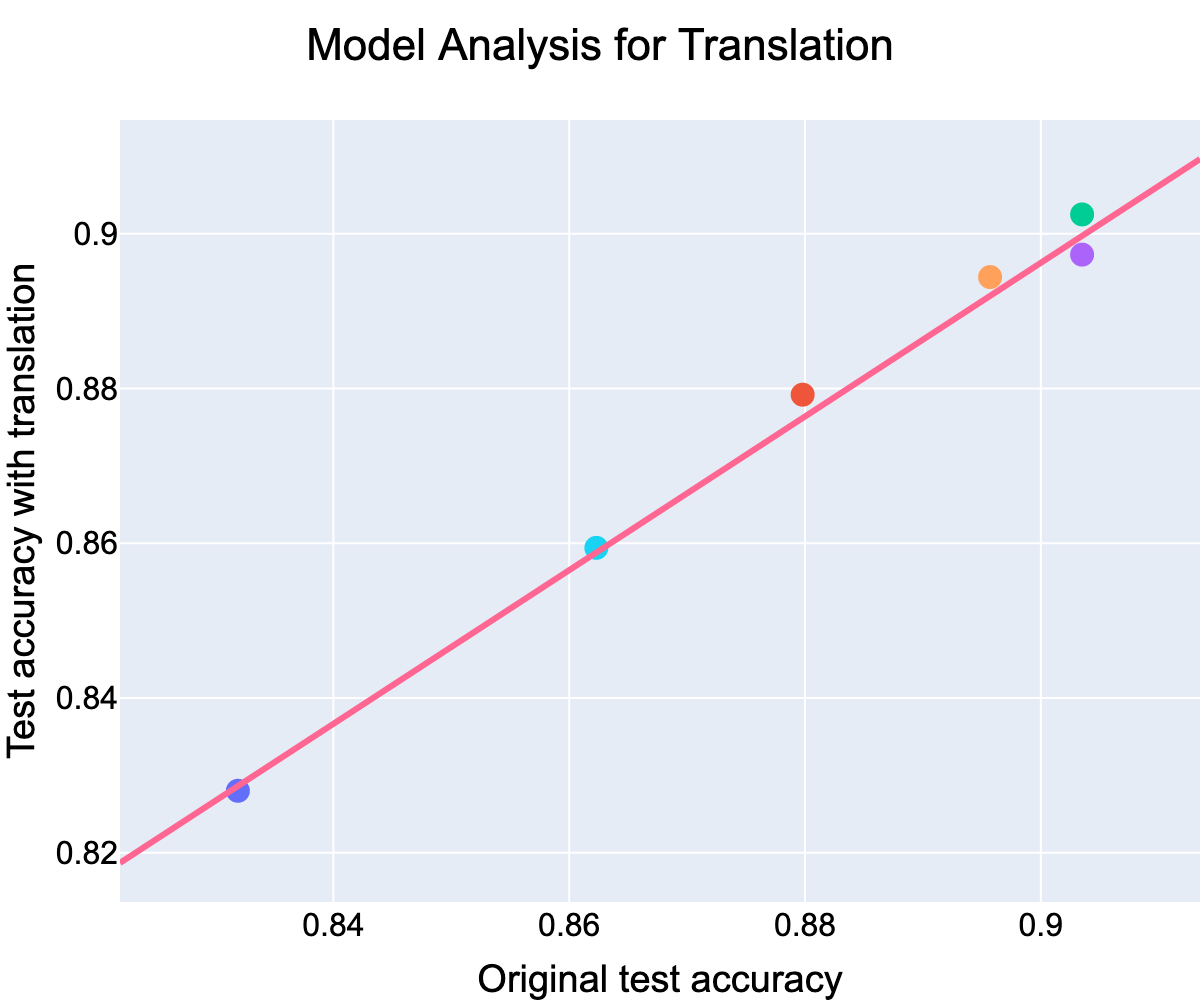}
  \end{subfigure}
  \caption{Translation test results. ViT shows comparable robustness \color{green}\checkmark}
  \label{fig:translation_test}
\end{figure*}

From 0 to 16 pixel translation, the test accuracy is measured (Figure \ref{fig:translation_test}). Both CNNs and ViTs showed good robustness for translation, and the drop in test accuracy was minor. We think that the weak translation invariance of ViT conjectured in the original paper is close to the functional property. As a statistical model, trained ViT has sufficient translation invariance.

Further, we explore the difference in robustness between ViTs and CNNs. We plotted the test accuracy of each model, which is trained on the Stanford Dogs dataset and tested at the center crop and 16-pixel translation. A trendline is drawn using the four points from CNNs. This trendline represents the expected test accuracy for the CNN family. If ViT exhibits a worse behavior than CNNs due to the difference in intrinsic architecture, ViT will locate under the trendline of CNNs. But ViT was near above CNNs' trendline, which means that ViT shows comparable or slightly improved translation invariance to CNNs. Additionally, for all architectures, the accuracy drop from translation appears only at a certain ratio, and the ranking between models hardly changed. This result shares the context with the study of \cite{recht2019imagenet}, who claimed that the accuracy drop was mainly caused by domain shift in the dataset. Therefore, we conclude that with respect to translation, ViTs have robustness comparable to CNNs.

\subsection{Other Tests}
\label{sec:othertests}
Similarly, the performance of CNNs and ViTs are measured in various corruptions including rotation, brightness, blur, gamma, saturation, and hue. In most cases, ViT showed performance comparable to CNNs or more robust (Figure \ref{fig:rotation_test}, \ref{fig:brightness_test}). See Appendix for the full results on various corruptions. However, some exceptions exist. For the brightness test, ViT was marginally weak. Further, ViT showed serious vulnerability to contrast enhancement.

\begin{figure*}
  \centering
  \begin{subfigure}{0.35\linewidth}
    \includegraphics[width=0.99\linewidth]{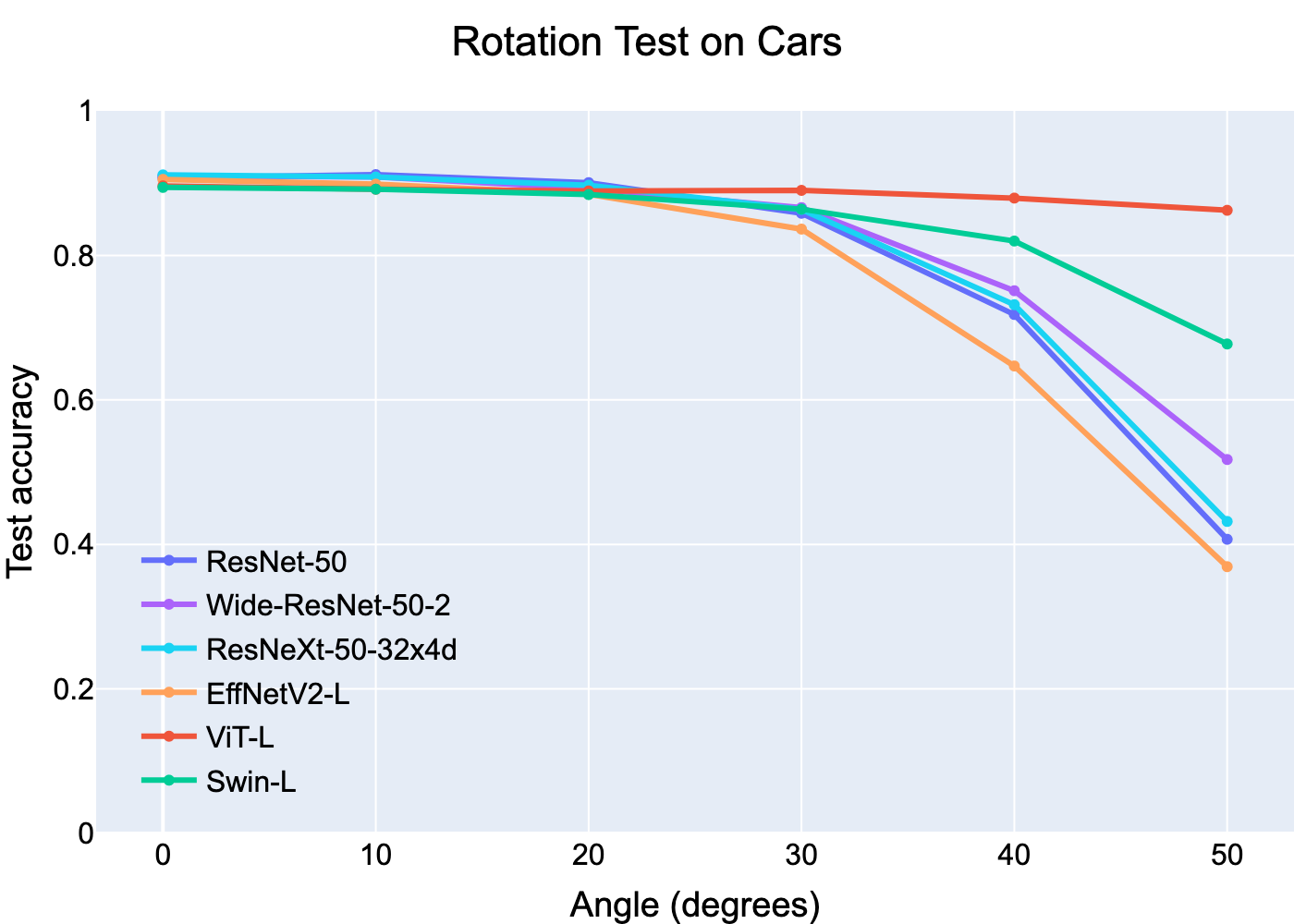}
  \end{subfigure}
  \hfill
  \begin{subfigure}{0.35\linewidth}
    \includegraphics[width=0.99\linewidth]{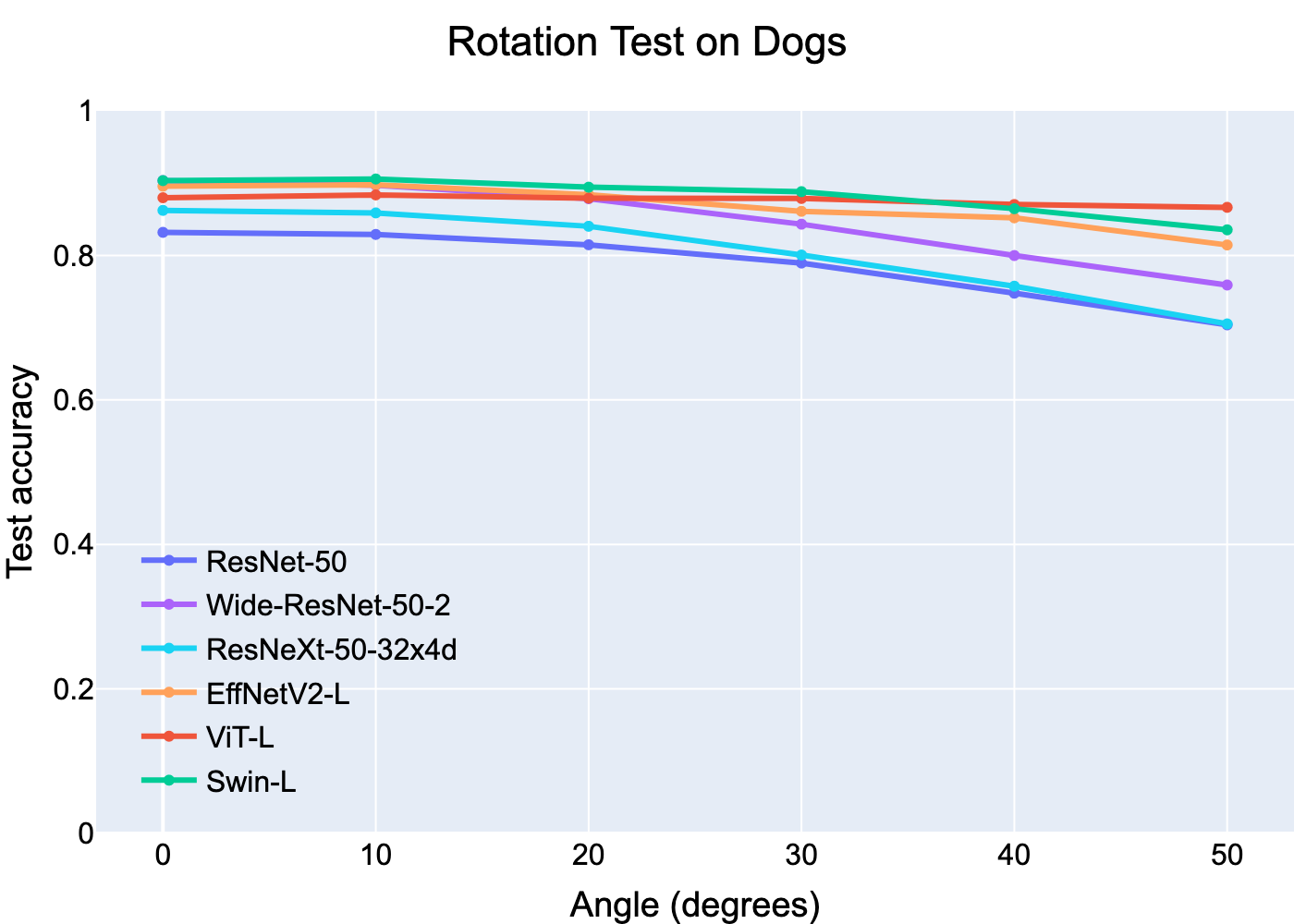}
  \end{subfigure}
  \hfill
  \begin{subfigure}{0.29\linewidth}
    \includegraphics[width=0.99\linewidth]{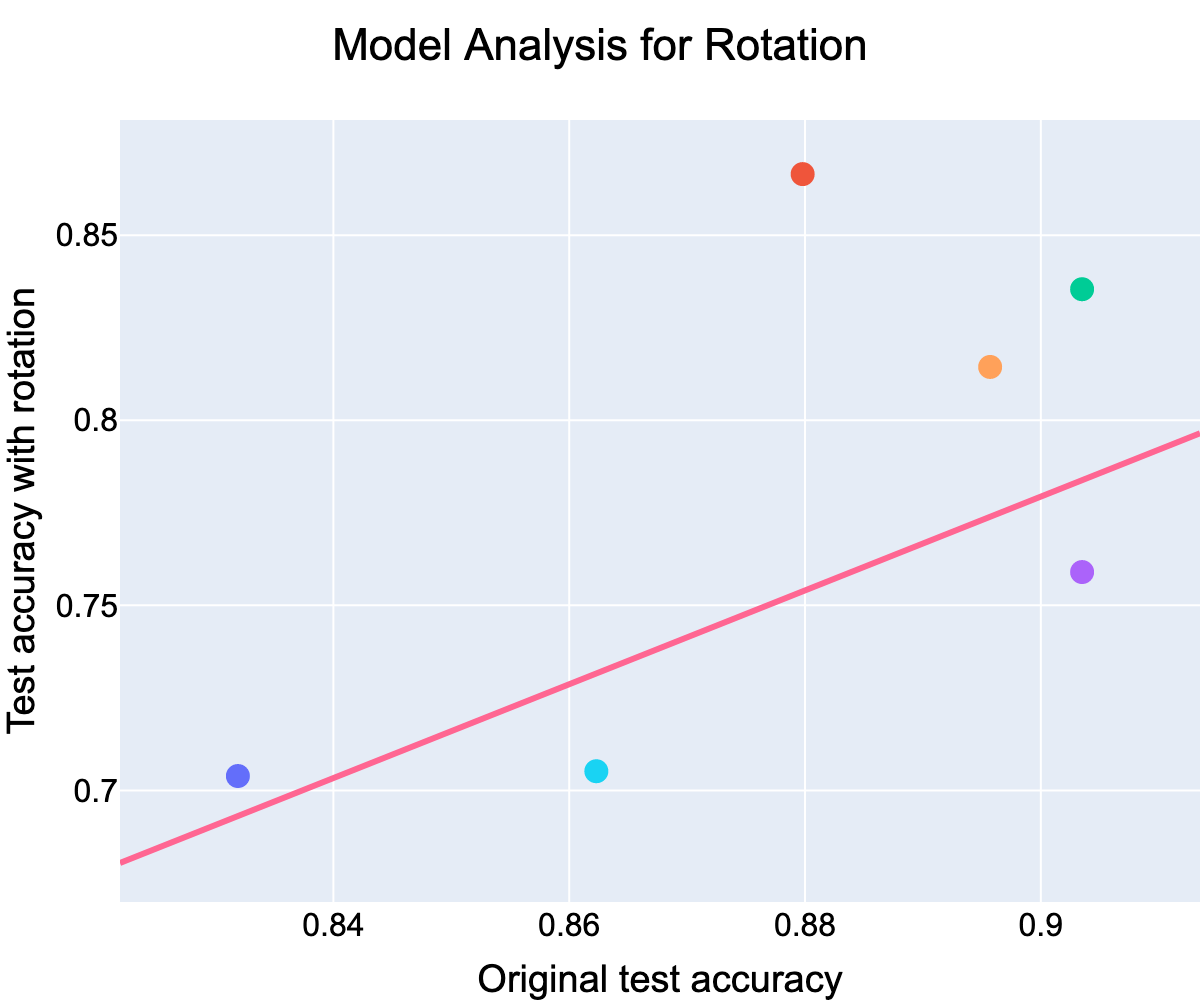}
  \end{subfigure}
  \caption{Rotation test results. ViT shows comparable robustness \color{green}\checkmark}
  \label{fig:rotation_test}
\end{figure*}

\subsection{Contrast Test}
\label{sec:contrasttest}

\begin{figure*}
  \centering
  \begin{subfigure}{0.35\linewidth}
    \includegraphics[width=0.99\linewidth]{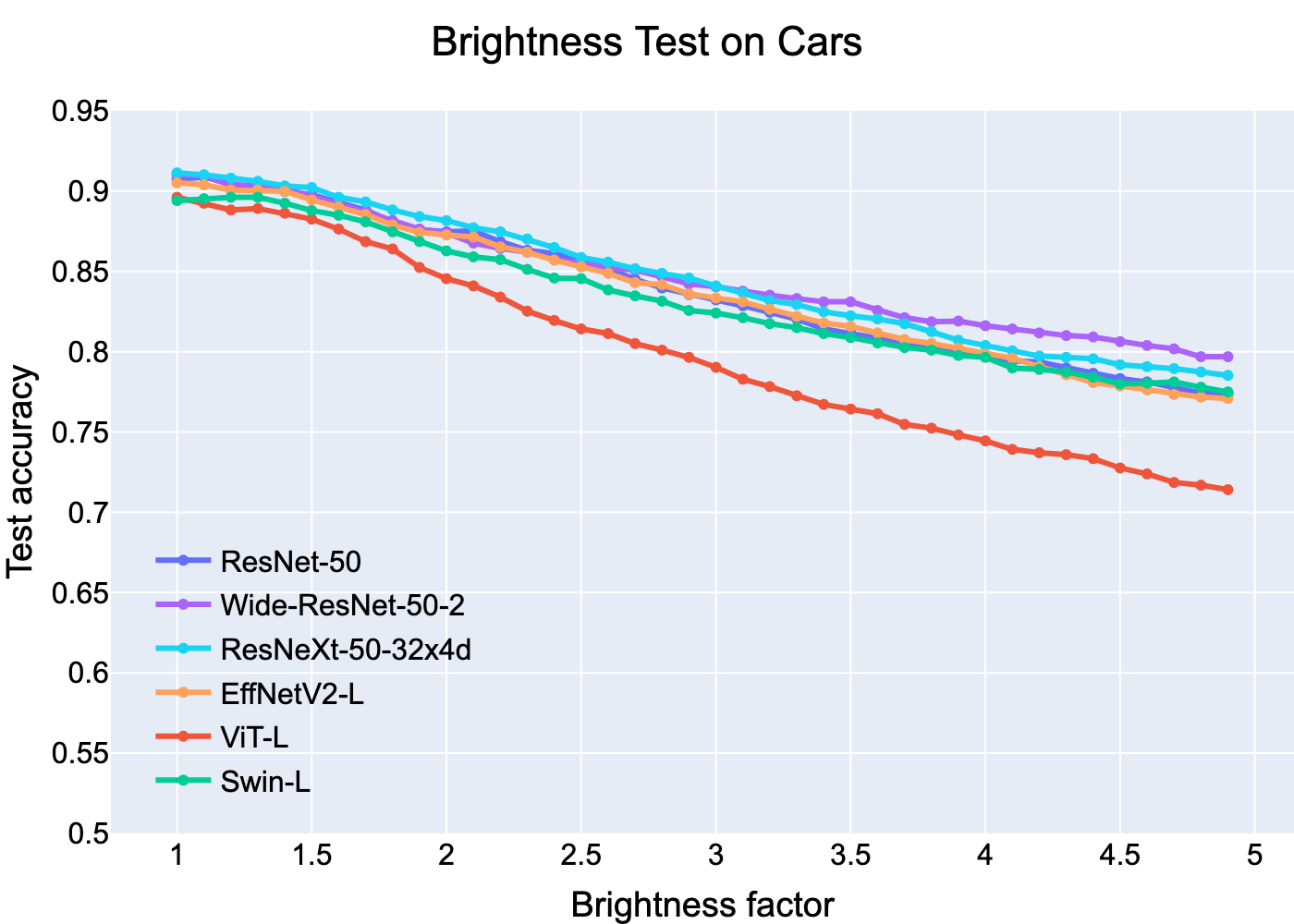}
  \end{subfigure}
  \hfill
  \begin{subfigure}{0.35\linewidth}
    \includegraphics[width=0.99\linewidth]{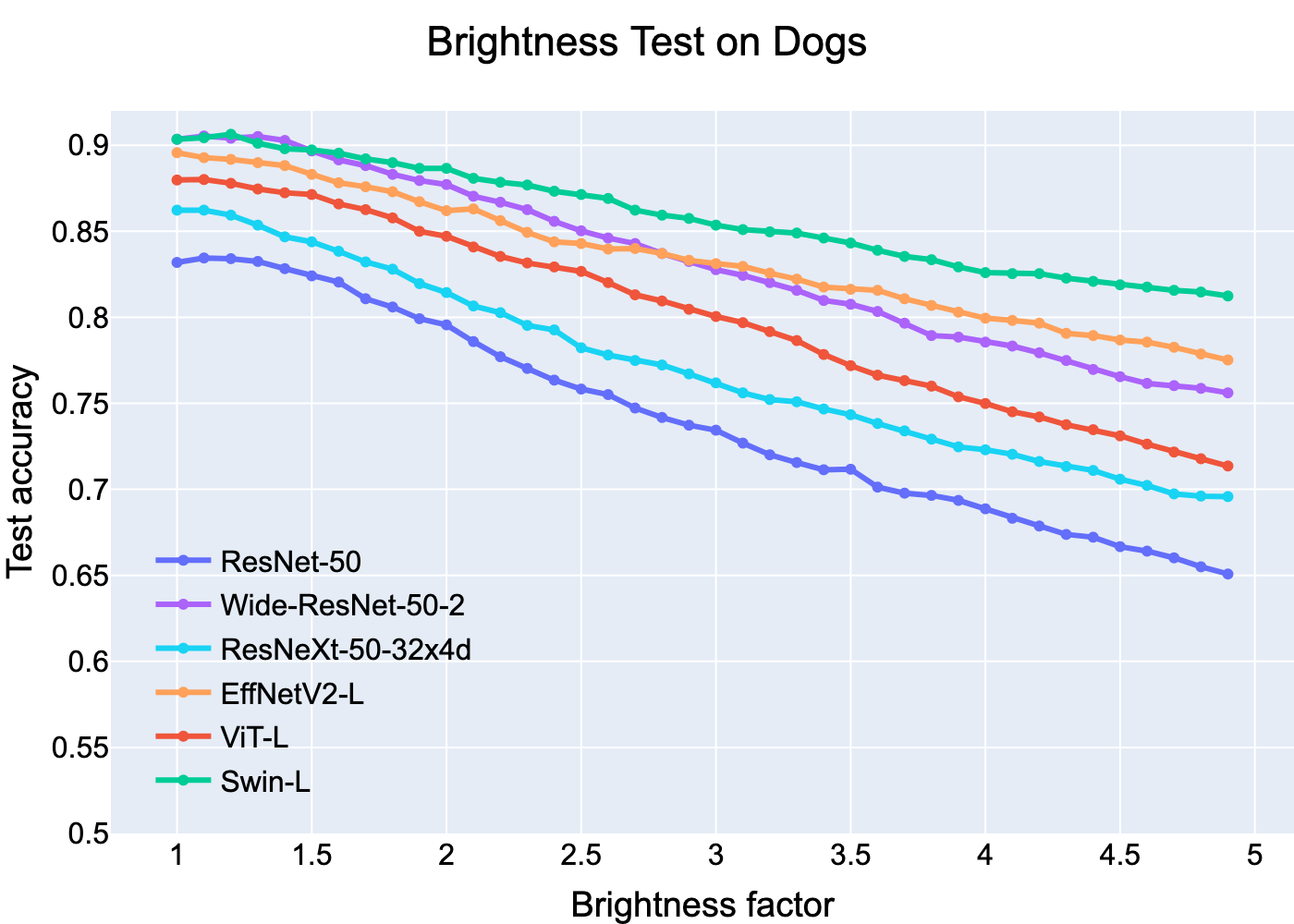}
  \end{subfigure}
  \hfill
  \begin{subfigure}{0.29\linewidth}
    \includegraphics[width=0.99\linewidth]{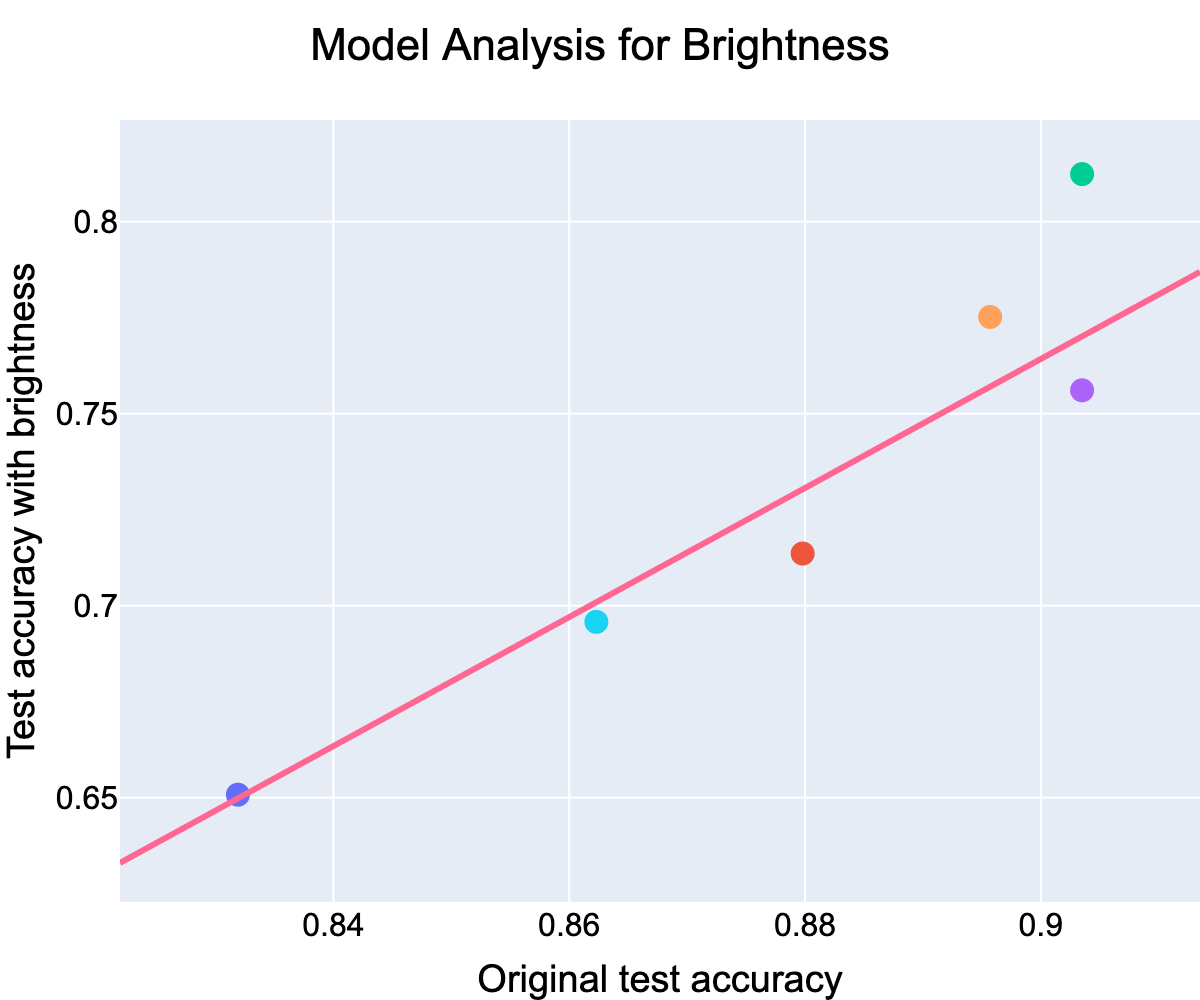}
  \end{subfigure}
  \caption{Brightness test results. ViT is vulnerable to brightness-varying environment \color{red}\xmark}
  \label{fig:brightness_test}
\end{figure*}

\begin{figure*}
  \centering
  \begin{subfigure}{0.35\linewidth}
    \includegraphics[width=0.99\linewidth]{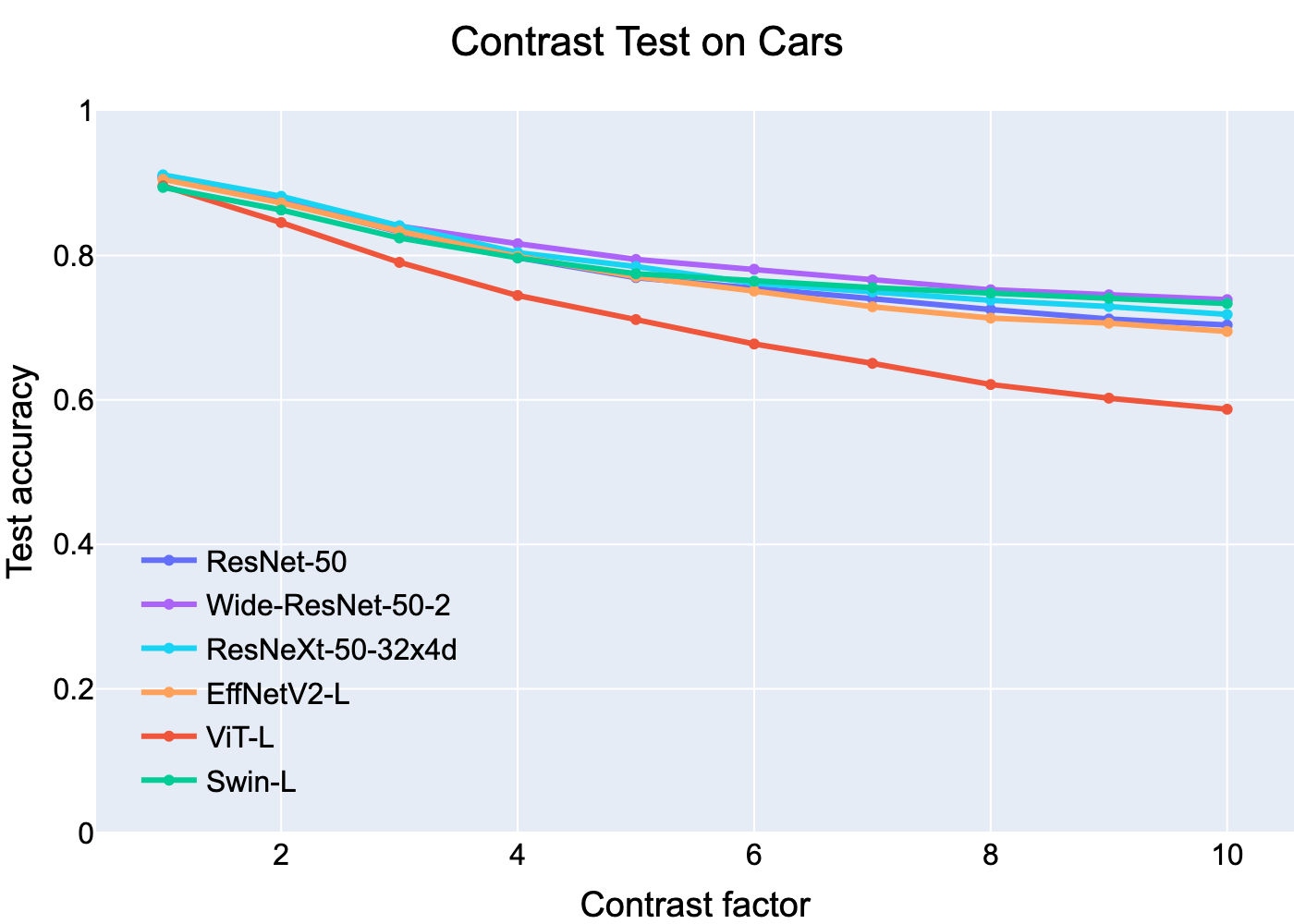}
  \end{subfigure}
  \hfill
  \begin{subfigure}{0.35\linewidth}
    \includegraphics[width=0.99\linewidth]{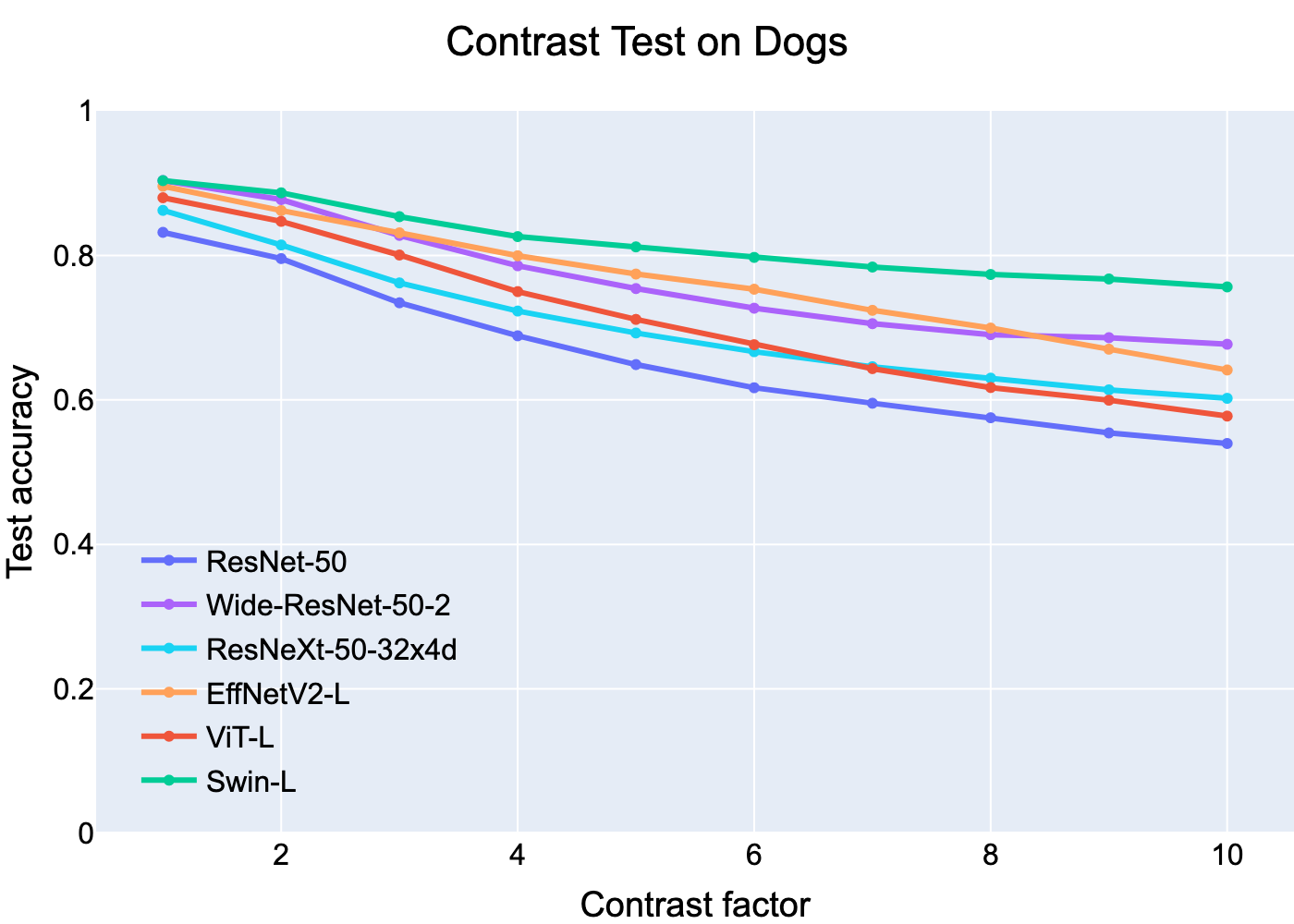}
  \end{subfigure}
  \hfill
  \begin{subfigure}{0.29\linewidth}
    \includegraphics[width=0.99\linewidth]{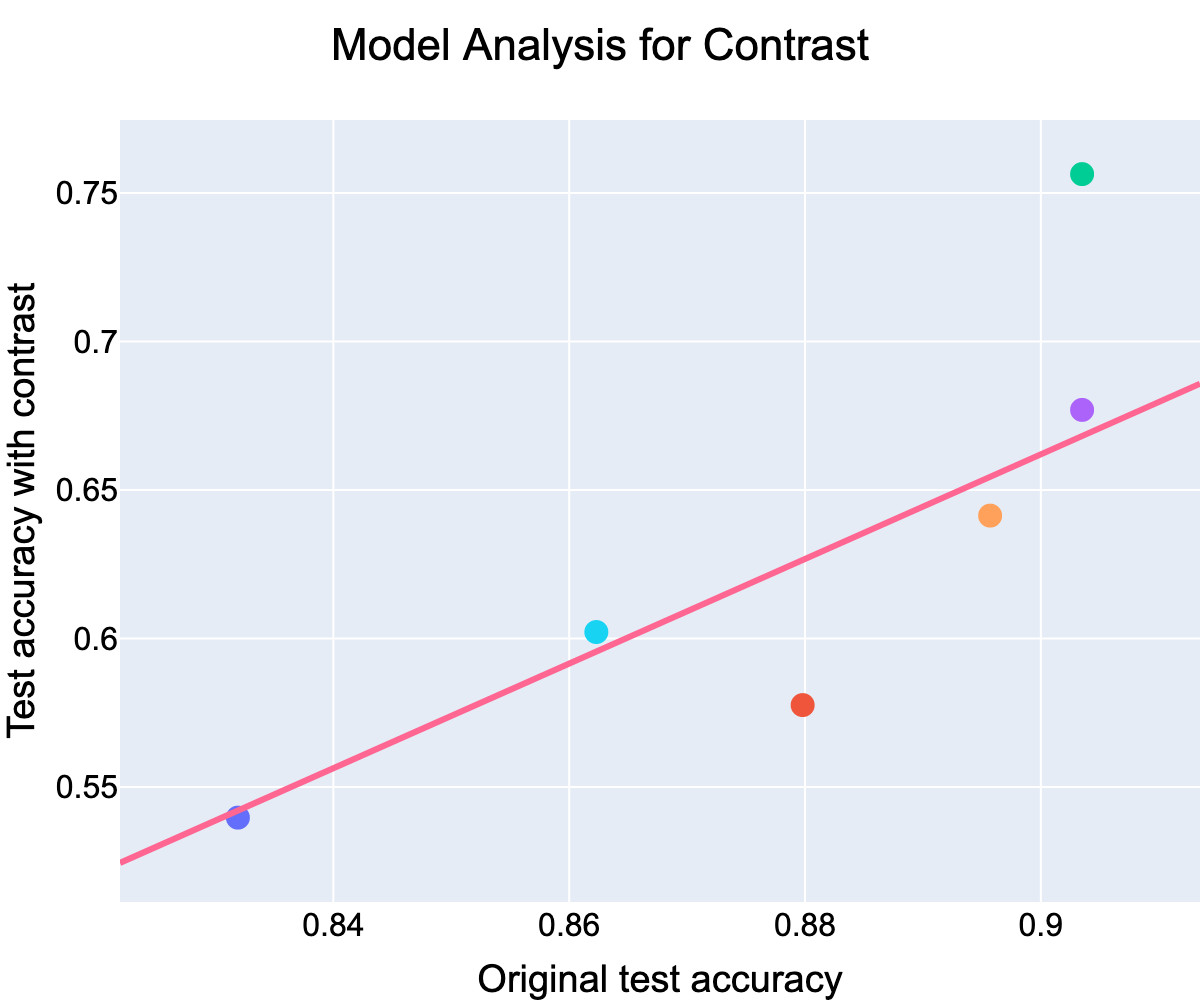}
  \end{subfigure}
  \caption{Contrast test results. ViT is vulnerable to contrast-varying environment \color{red}\xmark}
  \label{fig:contrast_test}
\end{figure*}

Contrast enhancement makes bright areas of the image brighter and dark areas darker. We observed that ViT is highly vulnerable to contrast-varying environments than CNNs (Figure \ref{fig:contrast_test}). This phenomenon was observed consistently with large margins for the two datasets. In contrast, Swin showed robust performance for contrast enhancement.

\section{Detailed Analysis on Patch Embedding}
\label{sec:detailedanalysisonpatchembedding}
Why is ViT vulnerable to contrast-varying environments? Meanwhile, why is Swin robust against contrast? From their architectural differences, we can consider various factors such as attention, patch embedding, model size and merging layer.

\begin{figure*}
  \centering
  \begin{subfigure}{0.49\linewidth}
    \includegraphics[width=0.99\linewidth]{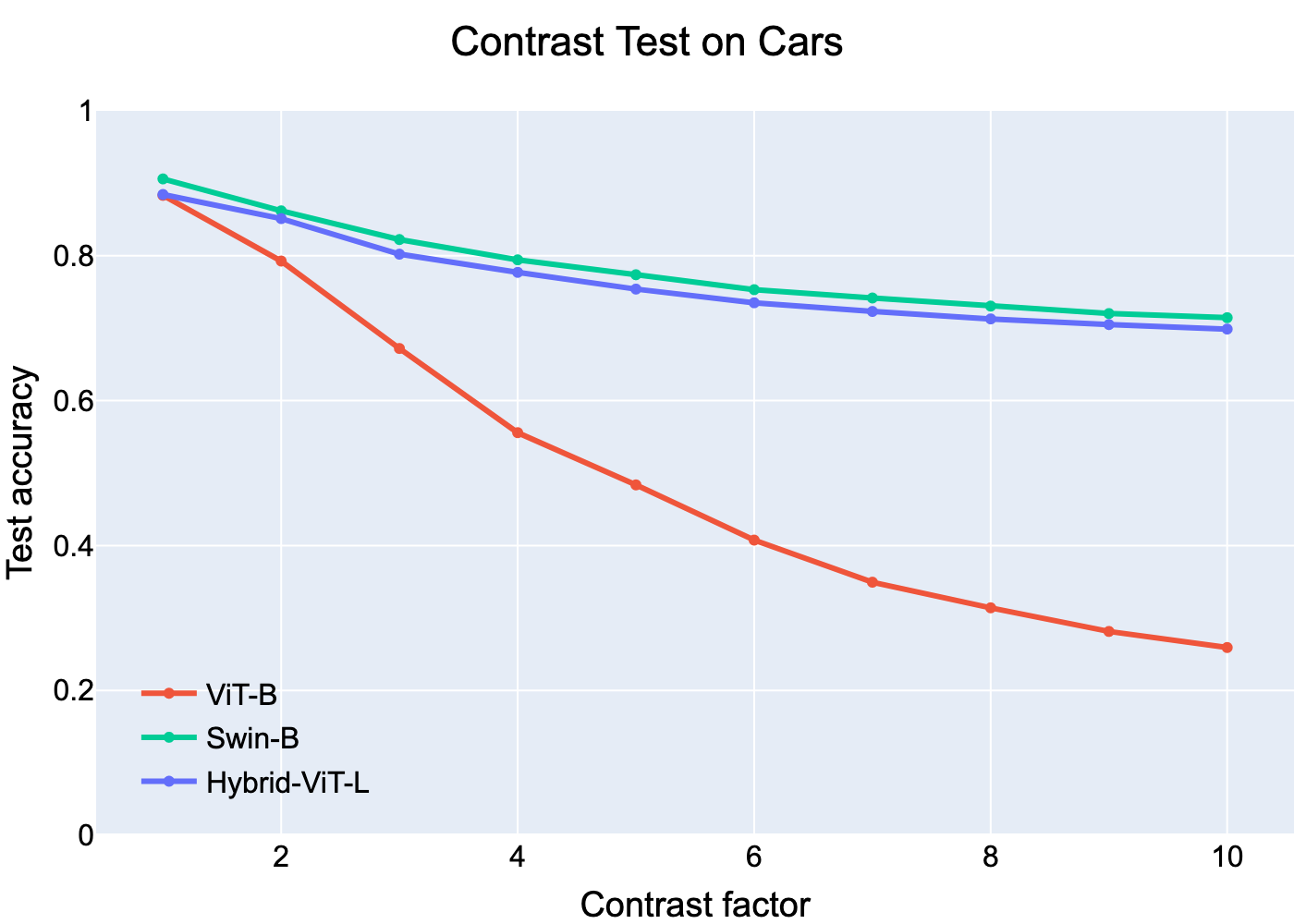}
  \end{subfigure}
  \hfill
  \begin{subfigure}{0.49\linewidth}
    \includegraphics[width=0.99\linewidth]{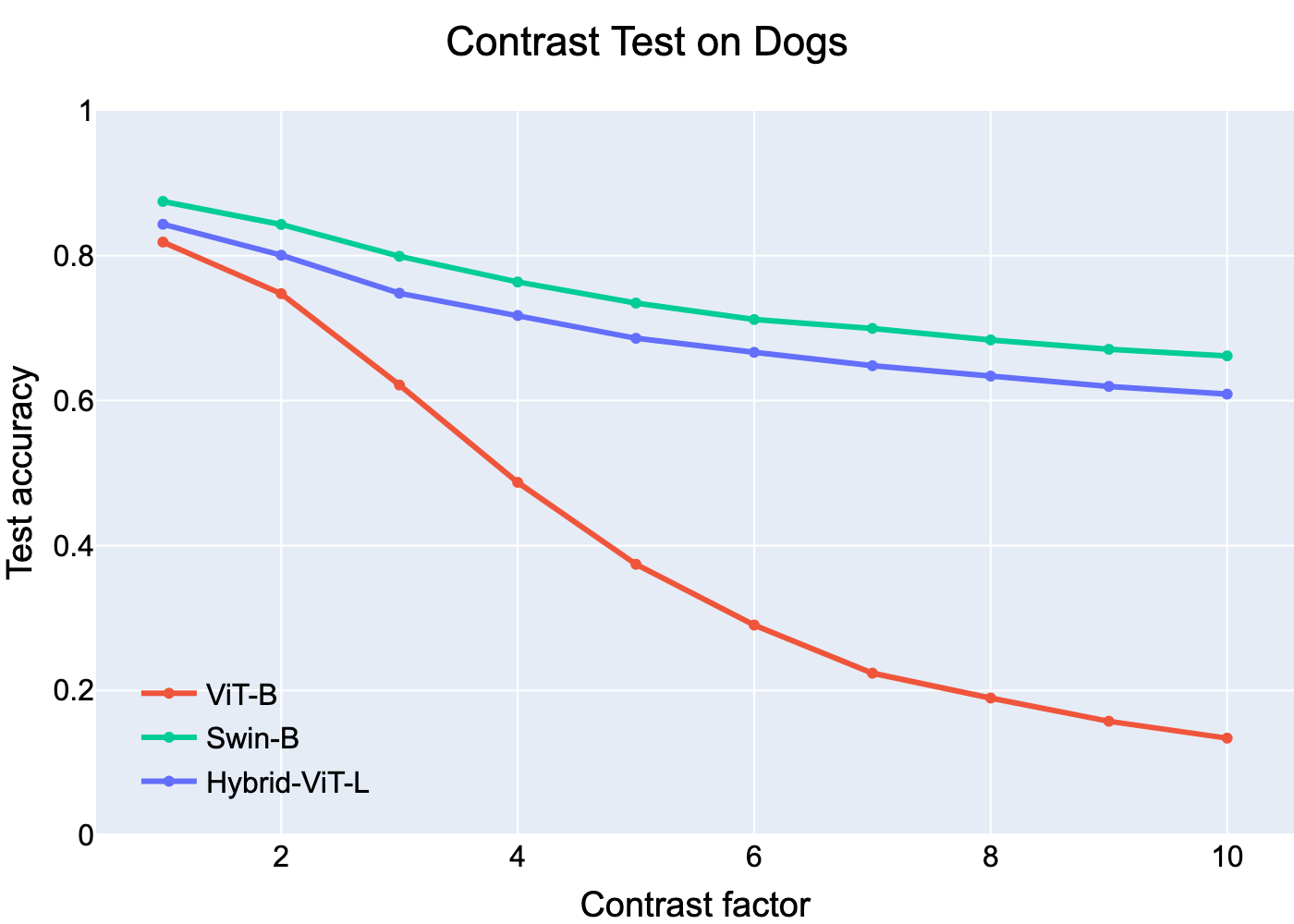}
  \end{subfigure}
  \caption{Contrast test on other ViT variants}
  \label{fig:contrast_test_additional}
\end{figure*}

Additionally, we performed the contrast test on ViT-B, Swin-B and Hybrid-ViT-L (Figure \ref{fig:contrast_test_additional}). First, ViT-B and Swin-B showed similar results with their large models. Meanwhile, interestingly, Hybrid-ViT-L showed robust properties for contrast enhancement. In the early stage, instead of dividing the image into patches, Hybrid-ViT employs ResNet's feature extractor, and in later stages, it uses the same self-attention and MLP blocks as ViT. Therefore, from the difference between ViT and Hybrid-ViT, the cause of ViT's contrast weakness seems to be related to the early stage configuration surrounding patch embedding.

\begin{figure}[t]
  \centering
   \includegraphics[width=0.99\linewidth]{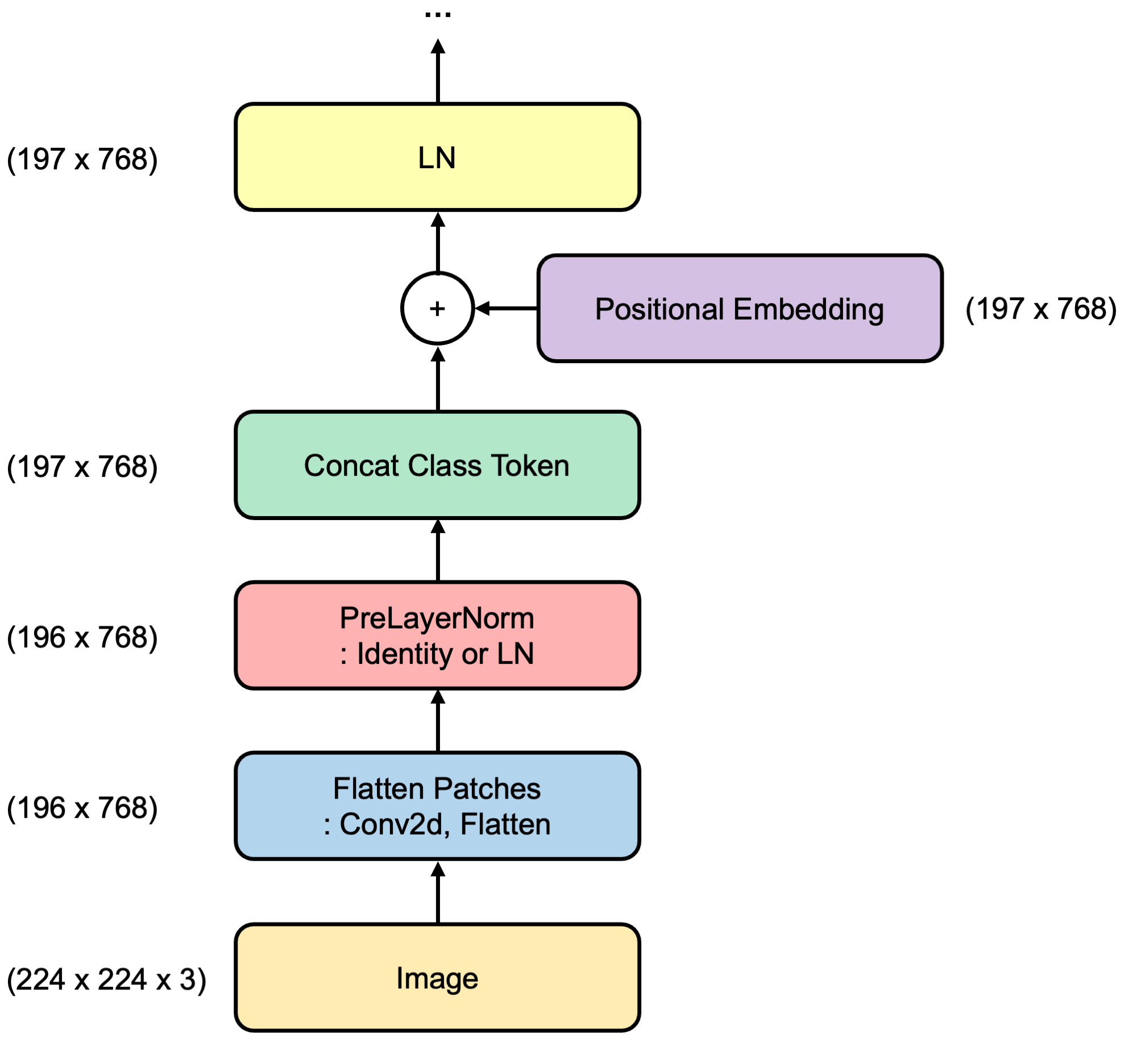}
   \caption{Illustation for early stage of ViT}
   \label{fig:early_stage}
\end{figure}

The early stages of ViT and Swin are shown in Figure \ref{fig:early_stage}. ViT first divides an image into patches using a strided convolution, then flatten them. Then, optional LayerNorm, which we call \textit{PreLayerNorm}, can be applied. ViT concatenates the class token here, which we ignore for simplicity. After that, the learnable positional embedding of the same size is added to reflect the positional difference like BERT \cite{devlin2018bert}. Now the Transformer Encoder starts with the LayerNorm, which we call \textit{PostLayerNorm}.

Here, we pay attention to PreLayerNorm and positional embedding. While ViT does not apply the optional LayerNorm, Swin employs LayerNorm. Thus, for flattened patches $X$, early-stage output up to PostLayerNorm is,
\begin{align}
z_{vit}(X) &= LN(X+E_{pos}), \label{eq:z_vit} \\
z_{swin}(X) &= LN(LN(X)+E_{pos}). \label{eq:z_swin}
\end{align}
Here, positional embedding behaves like a fixed bias during testing. However, in ViT, the scale of operand $X$ can be changed according to the image. Since the sum $X+E_{pos}$ is normalized by the subsequent PostLayerNorm, the effect of adding positional embedding appears relative to the scale of operand $X$. For example, if the scale of operand $X$ increases, the effect of adding $E_{pos}$ would vanish. Likewise, in ViT, as the scale of $X$ changes, the effect of positional embedding shows inconsistency.

In contrast, Swin employs PreLayerNorm. Even if the scale of $X$ is changed, the scale of $LN(X)$ is fixed. In other words, when PreLayerNorm is applied, since the scale of operand $LN(X)$ is guaranteed to be fixed, the effect of adding the positional embedding exhibits consistency.

Let's look at this phenomenon more mathematically. All LayerNorms \cite{ba2016layer} in the ViT family are applied to the channel dimension (e.g., 768), and unlike BatchNorm \cite{ioffe2015batch}, input statistics are always used in the train/test phase. We express LayerNorm as $LN(X)=L(N(X))$, where normalization $N(\cdot)$ step results in a normal distribution using the mean and variance of the input and linear transform $L(\cdot)$ step rescales the distribution with $\gamma$ and $\beta$. Note that the flattened patches $X$ result from the linear operation on the image. Thus, the flattened patches by the image to which per-channel scaling and bias are applied can be expressed as $aX+b$. Here, when PreLayerNorm is applied, since scaling and bias are reflected in the mean and variance of $N(\cdot)$ step, it is eventually normalized to the same distribution as the original. In other words,
\begin{align}
    N_{pre}(aX+b) = N_{pre}(X).
\end{align}
Thus, with PreLayerNorm, $LN_{pre}(aX+b)=LN_{pre}(X)$. Therefore, for the early stage output $z_{swin}(X)=LN(LN(X)+E_{pos})$,
\begin{align}
    z_{swin}(aX+b)=z_{swin}(X).
\end{align}
This means that when PreLayerNorm is applied, the early stage output works consistently even if a change in scale and bias in the image appears.

\begin{figure*}[ht!]
  \centering
  \begin{subfigure}{0.49\linewidth}
    \includegraphics[width=0.99\linewidth]{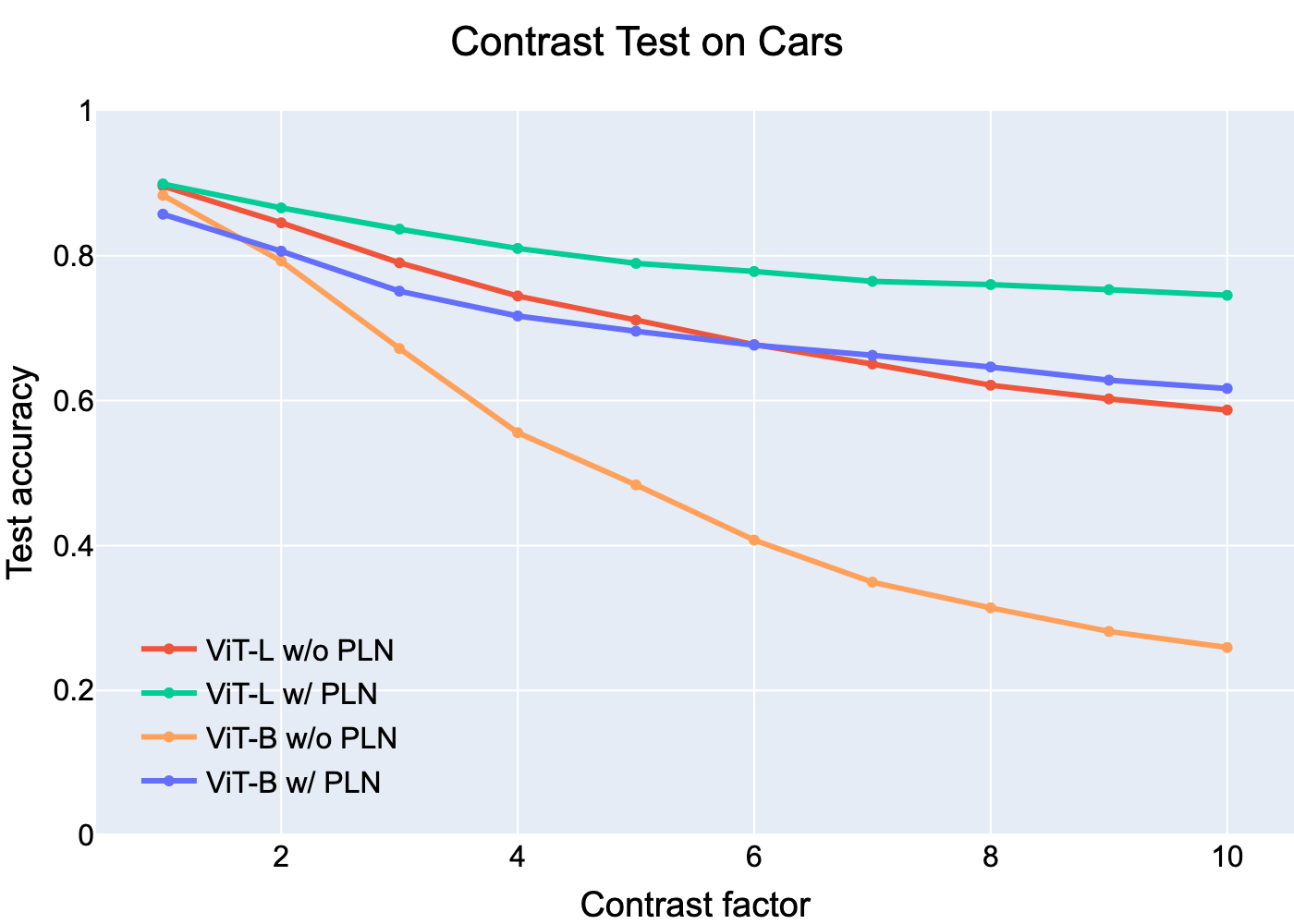}
  \end{subfigure}
  \hfill
  \begin{subfigure}{0.49\linewidth}
    \includegraphics[width=0.99\linewidth]{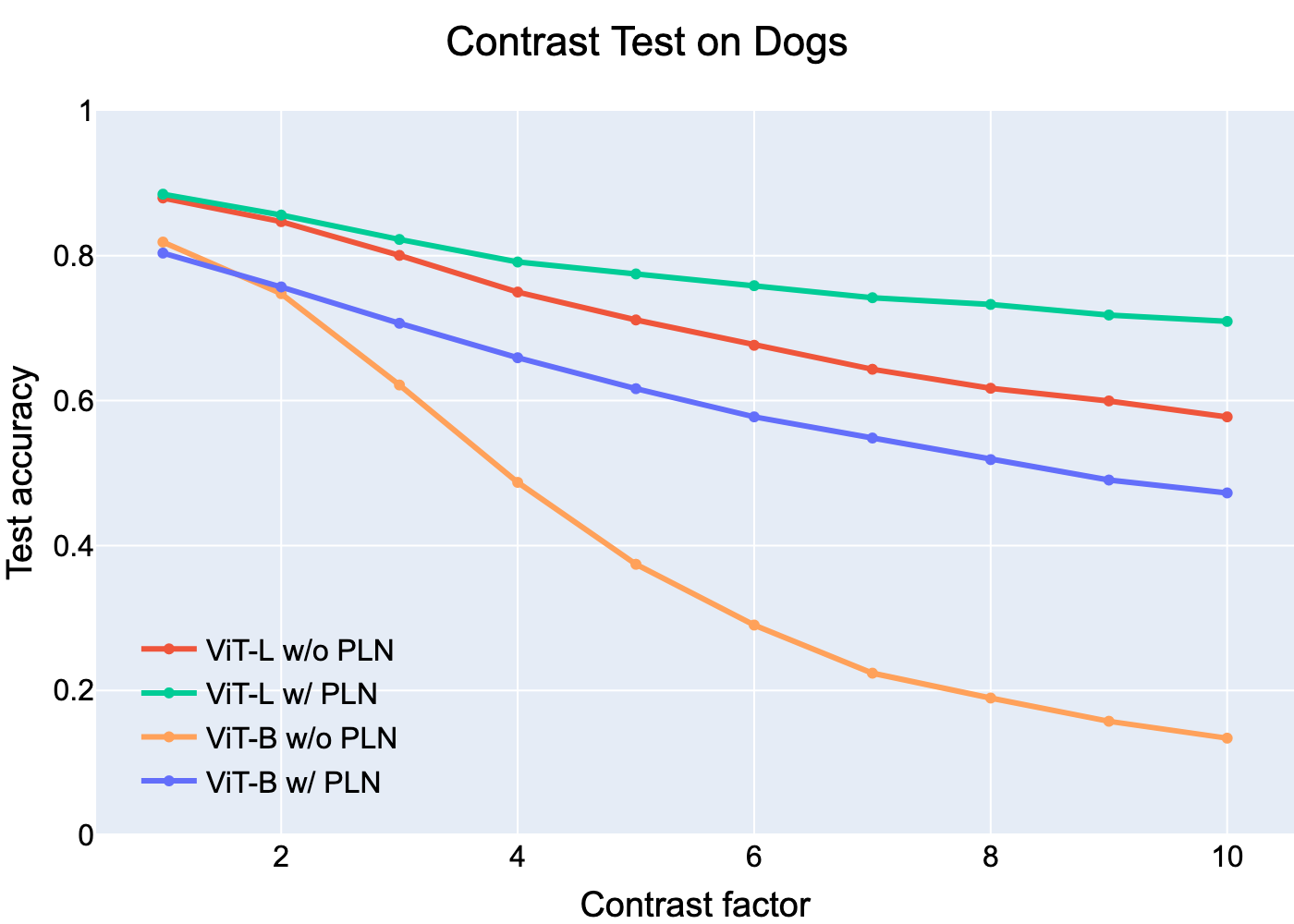}
  \end{subfigure}
  \caption{Contrast test results. ViT with PreLayerNorm shows improved robustness \color{green}\checkmark}
  \label{fig:contrast_test_pln}
\end{figure*}

In contrast, if there is no PreLayerNorm like ViT, the effect of adding positional embedding shows inconsistency depending on the scale and bias of the image.
\begin{theorem}
\label{theorem_1}
Assume a patch embedding without PreLayerNorm, where the early stage output is $z_{vit}(X)=LN(X+E_{pos})$. Since $N_{post}(aX+b+E_{pos}) \neq N_{post}(X+E_{pos})$, $z_{vit}(X)$ exhibits inconsistency with scale and bias, i.e.,
\begin{align}
z_{vit}(aX+b) \neq z_{vit}(X). \label{eq:vit_inconsistency}
\end{align}
\end{theorem}
See Appendix for the detailed proof. We claim that ViT has a potential problem in that the effect of adding positional embedding becomes inconsistent when the scale and bias of an image are changed. This phenomenon is presumed to be the cause of ViT's weakness in contrast enhancement. Therefore, ViT must employ PreLayerNorm like Swin to operate robustly in contrast-varying environments.

Finally, for ViT, we applied PreLayerNorm, performed fine-tuning, and then performed the same contrast test (Figure \ref{fig:contrast_test_pln}). Now, ViT with PreLayerNorm showed robustness comparable to CNNs and Swin for contrast enhancement.

\section{Empirical Observation of Vanishing Positional Embedding}
\label{sec:empiricalobservationofvanishingpositionalembedding}
As described above, the effect of positional embedding on the early stage output $z$ would change depending on the color scale of the image. To measure the effect of positional embedding on $z$, we propose an index, called \textit{effective contribution of positional embedding (ECPE)}. The contribution of positional embedding can be investigated through gradient.

Let a $i$th single image $I_i$ in dataset passes through ViT. First, the resulting $z \in R^{N \times D}$ are aggregated to be a scalar $Z=\sum_{n,d} z_{n,d}$. Then we examine the gradient of $Z$ with respect to positional embedding to obtain the amount of contribution of positional embedding to $Z$. However, since a single image does not represent general property, to obtain the average behavior of positional embedding on the dataset, gradients are accumulated over a large number of images. Meanwhile, since negative values in gradients cancel out positive values, we would like to ignore the negative importance. Therefore, after applying $ReLU$ to each gradient, we accumulate them. In short, the effective contribution of positional embedding is,
\begin{align}
    ECPE = \sum_{i} \sum_{n,d} ReLU\Big(\frac{\partial Z}{\partial E_{pos}}\Big).
\end{align}
ECPE represents the general contribution of positional embedding to $z$. Our claims are that 1) without PreLayerNorm, when the color scale of the image is changed, ECPE appears differently, representing inconsistent behavior of ViT, and 2) with PreLayerNorm, even when the scale of the image changes, ECPE and ViT's behavior show consistency.

\begin{figure*}
  \centering
  \begin{subfigure}{0.22\linewidth}
    \includegraphics[width=0.99\linewidth]{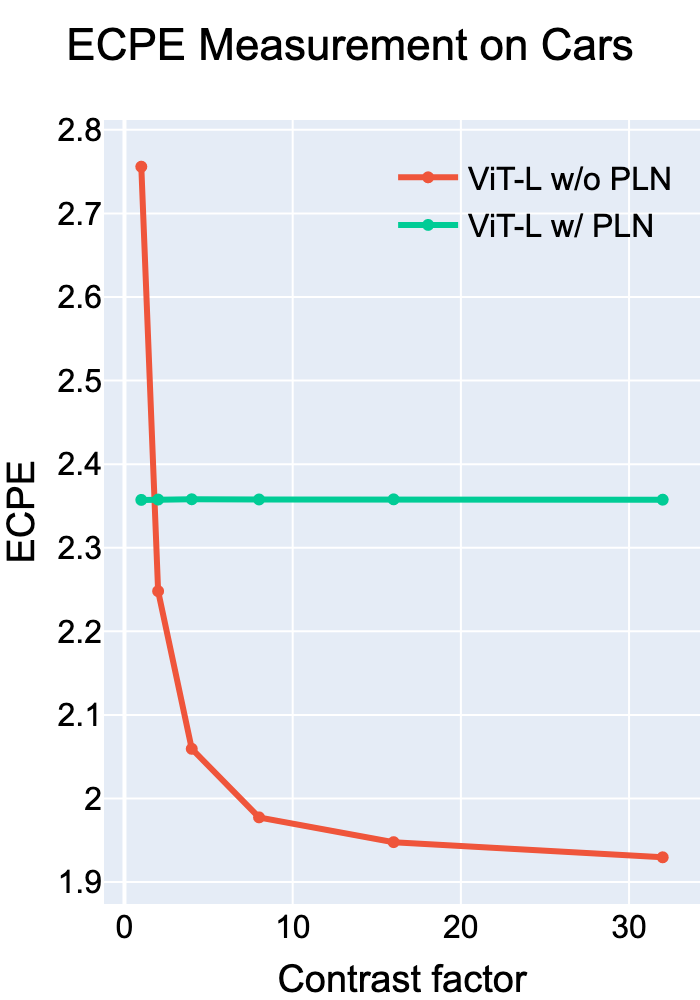}
  \end{subfigure}
  \hfill
  \begin{subfigure}{0.22\linewidth}
    \includegraphics[width=0.99\linewidth]{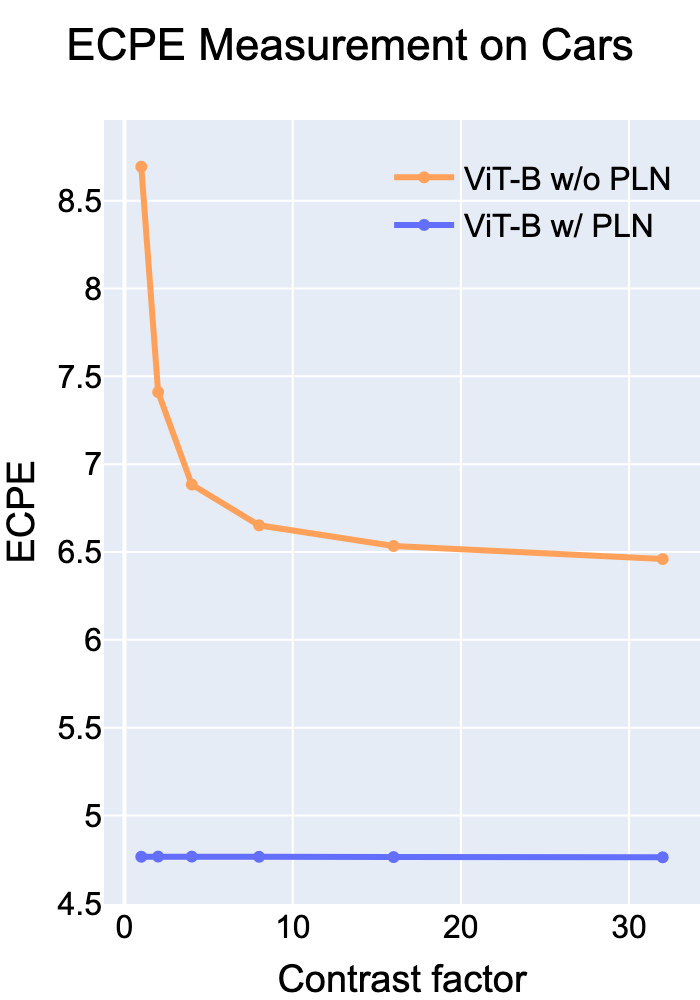}
  \end{subfigure}
  \hfill
  \begin{subfigure}{0.22\linewidth}
    \includegraphics[width=0.99\linewidth]{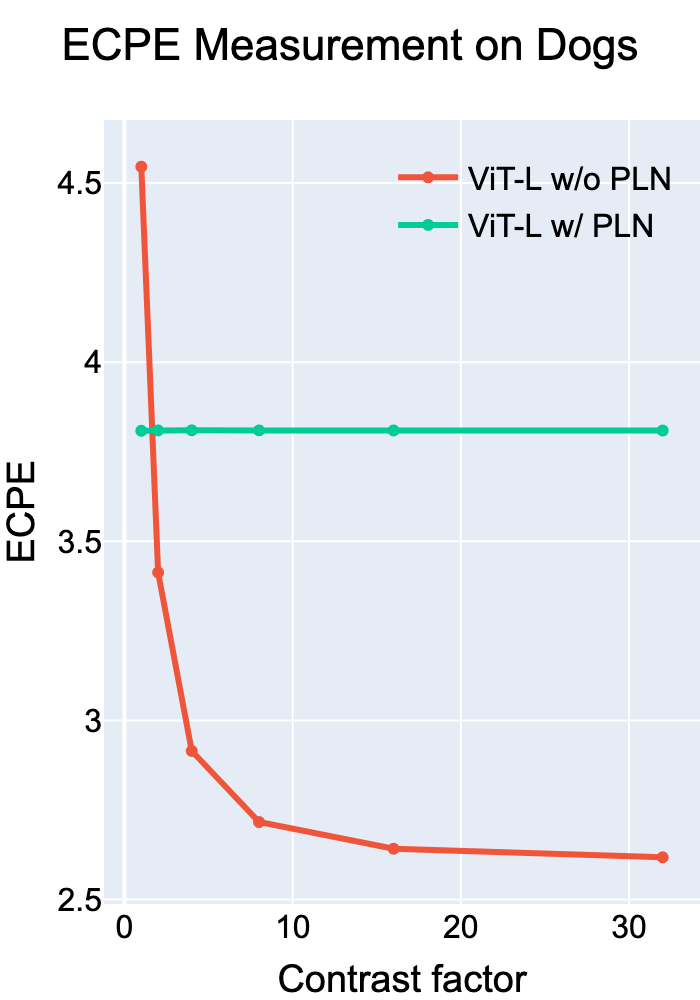}
  \end{subfigure}
  \hfill
  \begin{subfigure}{0.22\linewidth}
    \includegraphics[width=0.99\linewidth]{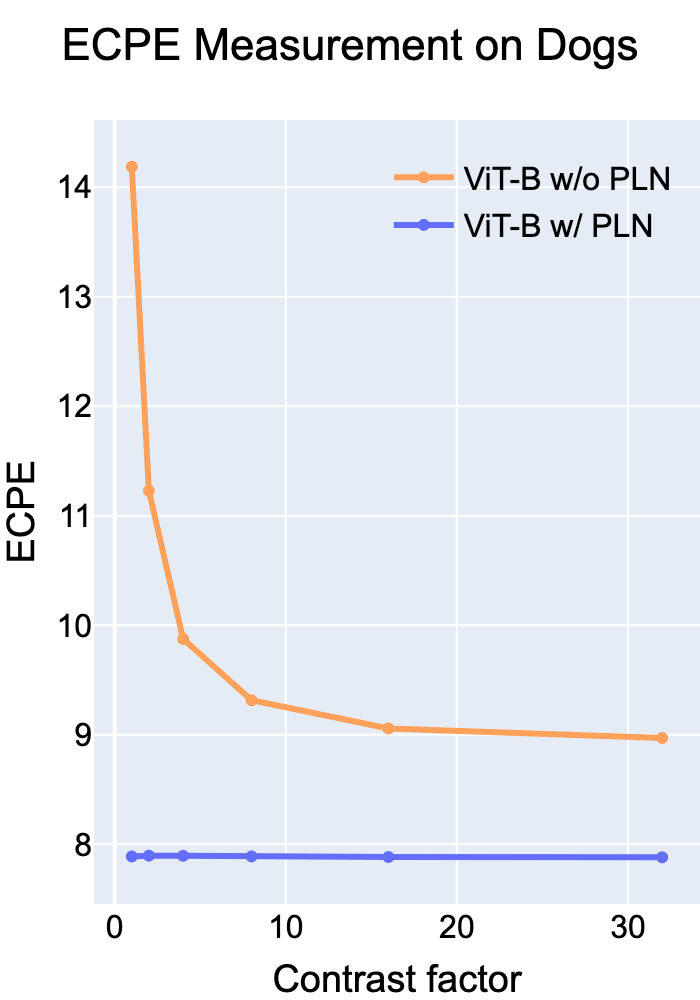}
  \end{subfigure}
  \caption{ECPE measurement in contrast-varying environment. PreLayerNorm enables the consistent behavior of positional embedding}
  \label{fig:ecpe}
\end{figure*}

We experimentally measured the ECPE in a contrast-varying environment. We targeted the Stanford Cars and Dogs datasets and the fine-tuned ViT models. ECPE was measured when each image was contrast-enhanced (Figure \ref{fig:ecpe}). First, without PreLayerNorm, ECPE decreased according to contrast enhancement factors. This observation is consistent with our claim that the effect of adding positional embedding becomes relatively small when the contrast factor is larger than 1. In other words, the effect of positional embedding vanishes when the color scale increases. In contrast, with PreLayerNorm, ECPE appeared as a consistent value even with contrast enhancement. This behavior of ECPE on the scale clearly demonstrates our claim that PreLayerNorm ensures a consistent effect of positional embedding.

%\begin{table}
%  \centering
%  \begin{tabular}{@{}lc@{}}
%    \toprule
%    Method & Frobnability \\
%    \midrule
%    Theirs & Frumpy \\
%    Yours & Frobbly \\
%    Ours & Makes one's heart Frob\\
%    \bottomrule
%  \end{tabular}
%  \caption{Results.   Ours is better.}
%  \label{tab:example}
%\end{table}

\section{Discussion}
\label{sec:discussion}
\subsection{Understanding Contrast Enhancement}
Brightness enhancement brightens all areas of the image, while contrast enhancement makes bright areas of the image brighter and dark areas darker. In the \texttt{Python Imaging Library (PIL)} implementation, both brightness and contrast enhancements take the form:
\begin{align}
    I_{out} = factor * I_{in} + degenerate.
\end{align}
Here, brightness enhancement uses a zero-valued image as degenerate, whereas contrast enhancement uses the channel mean as degenerate. If bias is ignored, the effect of color scale change through brightness and contrast enhancement is the same. Therefore, ideally, if this enhancement is applied with a specific factor, the flattened patches undergo the same color scaling.

However, the representation of images is constrained to RGB integers from 0 to 255. For example, when brightness enhancement is applied with a factor of 5, from the range of 0 to 255, only the range of 0 to 51 is linearly scaled, and the other range is saturated to 255. In other words, brightness enhancement becomes a non-linear scale transform for a bright area and cannot be expressed by a linear scale transform of $aX+b$. In order for brightness enhancement to be linear scaling, a factor around 1 should be used to ensure little saturation.

Meanwhile, contrast enhancement uses degenerate as the channel mean to brighten bright areas and darken dark areas. Therefore, saturation appears in both bright and dark areas, and the mid-range is linearly scaled. In this regard, we conjecture that contrast enhancement will be relatively closer to linear scale transform of $aX+b$. For this reason, the weakness of ViT will be observed more in contrast enhancement than brightness.

\subsection{Comparison with DeiT}
Though we focused on the differences in intrinsic architecture, we should discuss the pre-trained weights. Interestingly, we observed that the robustness of the pre-trained model affects the robustness of the fine-tuned model.

This phenomenon was confirmed in fine-tuning experiments using pre-trained weight from DeiT \cite{touvron2021training}. The architecture of non-distilled DeiT is the same as that of ViT. Unlike the large dataset of JFT-300 used in ViT, DeiT was trained on ImageNet using strong data augmentations and regularizations \cite{cubuk2020randaugment,zhang2017mixup,yun2019cutmix}. Here, we observed that the model fine-tuned from DeiT is more robust in contrast enhancement than ViT (See Appendix). We conjecture that a model trained with strong data augmentations and regularizations like DeiT will acquire stronger robustness, which will be helpful for the robustness of a fine-tuned model. However, since data augmentation and regularization cannot solve the intrinsic architectural problem we discussed, they will only be a partial solution.\footnote{Work in Progress}\\

%\section{Conclusion}
%\label{sec:conclusion}
%dd

\clearpage

\bibliographystyle{named}
\bibliography{egbib}

\end{document}